\documentclass{article}

\usepackage{arxiv}

\usepackage[utf8]{inputenc} 
\usepackage[T1]{fontenc}    
\usepackage{hyperref}       
\usepackage{url}            
\usepackage[normalem]{ulem}
\usepackage{multicol}        
\usepackage{booktabs}       
\usepackage{amsfonts}       
\usepackage{nicefrac}       
\usepackage{microtype}      
\usepackage{lipsum}		
\usepackage{graphicx}
\usepackage{natbib}
\usepackage{doi}

\title{Evaluating Splitting Approaches in the Context of Student Dropout Prediction}

\author{\href{https://orcid.org/0000-0001-7563-1139}{\includegraphics[scale=0.06]{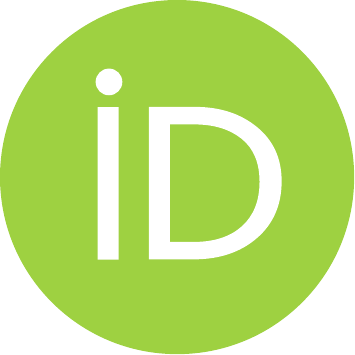}\hspace{1mm}Bruno de M. Barros} \\
	Informatics Institute\\
	Federal University of Goiás\\
	Goiânia GO, Brazil \\
	\texttt{brunomattos@discente.ufg.br} \\
	\And
	\href{https://orcid.org/0000-0003-1690-1201}{\includegraphics[scale=0.06]{orcid.pdf}\hspace{1mm}Hugo A. D. do Nascimento} \\
	Informatics Institute\\
	Federal University of Goiás\\
	Goiânia GO, Brazil \\
	\texttt{hadn@inf.ufg.br} \\
        \And
	\href{https://orcid.org/0000-0001-8500-4112}{\includegraphics[scale=0.06]{orcid.pdf}\hspace{1mm}Raphael Guedes} \\
	Informatics Institute\\
	Federal University of Goiás\\
	Goiânia GO, Brazil \\
	\texttt{braphaelguedes@egresso.ufg.br} \\
        \And
	\href{https://orcid.org/0000-0002-2155-012X}{\includegraphics[scale=0.06]{orcid.pdf}\hspace{1mm}Sandro E. Monsueto} \\
	Faculty of Administration, Accounting \\ Sciences and Economic Sciences\\
	Federal University of Goiás\\
	Goiânia GO, Brazil \\
	\texttt{monsueto@ufg.br} \\
}

\date{}

\renewcommand{\shorttitle}{Evaluating Splitting Approaches in the Context of Student Dropout Prediction}

\hypersetup{
pdftitle={A template for the arxiv style},
pdfsubject={q-bio.NC, q-bio.QM},
pdfauthor={David S.~Hippocampus, Elias D.~Striatum},
pdfkeywords={First keyword, Second keyword, More},
}

\begin{document}
\maketitle

\begin{abstract}
The prediction of academic dropout, with the aim of preventing it, is one of the current challenges of higher education institutions. Machine learning techniques are a great ally in this task. However, attention is needed in the way that academic data are used by such methods, so that it reflects the reality of the prediction problem under study and allows achieving good results. In this paper, we study strategies for splitting and using academic data in order to create training and testing sets. Through a conceptual analysis and experiments with data from a public higher education institution, we show that a random proportional data splitting, and even a simple temporal splitting are not suitable for dropout prediction. The study indicates that a temporal splitting combined with a time-based selection of the students' incremental academic histories leads to the best strategy for the problem in question.
\end{abstract}

\keywords{student dropout \and higher education \and educational data mining \and train/test split}

\section{Introduction}
The retention and dropout problems have been analyzed in several educational levels and formats. According to Lykourentzou et al.~\cite{Lykourentzou_et_al_2009}, these are among the main indicators that universities and policymakers use to evaluate the quality of an educational institution or program. Dropout indicators have been observed by international institutions such as The Organization for Economic Cooperation and Development (OECD), which finds that, globally, only 39\% of the students that enter the higher education system conclude in the proper time~\cite{OECD_2019}. From an economic point of view, academic dropouts create vacancies that cannot be easily recovered in a short period of time, generating under-utilization of human and financial resources in educational institutions~\cite{Yorke_2000}. Furthermore, dropout can have a significant negative emotional impact on students and on their families, discouraging further investments in human and social capital~\cite{Bonifro_et_al_2020}.

Therefore, a correct identification of dropout-prone students can help instructors and institutions to better address specific actions to reduce their probability to drop the course. However, the causes are diverse and can reside in the socioeconomic sphere or in the student's relationship with the academic environment, comprising performance, change of culture, among other aspects~\cite{Bardach_et_al_2020, Schnettler2020}. Since this is a highly complex phenomenon with many factors involved, predicting and designing comprehensive policies is also a challenging task. 

In this context, Machine Learning (ML) has become a great ally. There is an increasing scientific production on the use of ML to educational data mining, with focus on the analysis of many educational challenges, like academic performance and dropout predictions. The literature in this area investigates the potential effectiveness of a wide spectrum of ML methods and extracted features. However, not much attention has been given to the way in which an educational database should be split into training and test sets for supporting the application of ML methods, even though this choice may affect their outcomes. Many scientific works do not mention the splitting approaches they used,  or apply traditional approaches to data segmentation, such as cross-validation and hold-out. Even some works that explicitly mention temporal aspects in their methods do not segment the dataset in the same temporal way for the training and testing processes. Only a few set of papers~\cite{Ren_Ning_Rangwala_2017,Chen_Johri_Rangwala_2018,Krauss_Merceron_Arbanowski_2019,Nguyen_Vo_2019,Borrella_Caballero_Cueto_2019}
used inherently time-dependent approaches. Based on them, it can be concluded that temporal splitting is useful for creating more efficient predictive models for real-world educational data mining. However, temporal dataset splitting  is not the main focus of discussion of those studies. Some of them do not even concern academic dropouts, while others deal with dropouts but investigate cases with very few and short time-windows. In conclusion, there is still a lack of a conceptual and experimental investigation comparing different data-splitting approaches for large datasets and long academic periods, in order to support dropout prediction in higher education institutions.

In view of this gap in the literature, the objective of the present paper is to investigate the dropout prediction problem and show that a temporal splitting approach is more suitable for this task in higher education. We note that the dropout prediction process depends on academic records that are essentially time-stamped data. In addition, the prediction problem to be solved has a temporal nature, since it consists of predicting dropout changes for enrolled students based on knowledge about students who have already graduated or dropped out the educational institution. In order to support our claims, we perform a comparison between a proportional splitting and several temporal splitting approaches. Their inherent meaning and implications to the accuracy of machine learning methods in a real scenario are evaluated. The expected end result is to provide better predictions for the dropout problem, still in time to allow preventive and/or corrective pedagogical and socioeconomic actions.

The remainder of this work is organized as follows: Section~\ref{sec:basics} formally introduces some basic concepts that are useful for understanding a dropout prediction process; Section~\ref{sec:splits} is the key part of the paper, since it presents six splitting approaches for the definition of training and test sets and analyzes their main characteristics; Section~\ref{sec:evaluation} describes the evaluation of these approaches in a real scenario; Finally, Section~\ref{sec:conclusion} draws our conclusions and presents ideas for future research.


\section{Basic Definitions}
\label{sec:basics}
In this section, we introduce some assumptions and formal definitions in order to support a clear understanding of the splitting approaches in the context of dropout prediction for higher education institutions. 

We consider a pipeline for choosing a suitable machine learning method for dropout prediction. It follows a common sequence of steps, starting with a  data-gathering and preprocessing stage and ending with the output of the best method found so far. We assume that the educational institution has a database with academic records including general and less variable information about all students (such as the student's name, his/hers date of birth, sex, degree name and degree enrollment data), as well as historical data regarding every single course taken by the students (e.g., the course name, the academic term of enrollment, the attendance percentage and the obtained score in the course). We also consider that the database has information about the last \textit{enrollment status} of each student, which, for simplicity, is referred here just as ``graduated'' (for those who successfully completed their degree), ``dropout'' (when the student left the institution without completing the degree) and ``enrolled'' (when the student is still doing the degree).  Furthermore, when a student leaves the educational institution (either by graduating or dropping out), his or her \textit{exit date} (or \textit{exit term}) must be registered in the database.  In the pipeline, the first two steps concern gathering these pieces of data and  preprocessing and grouping them in order to create a new data structure for every student with his/her descriptive features and academic status. During this process, inconsistent and incomplete data are treated and new pieces of information  (for example, statistical information) are produced and added to the student structure. Usually, feature extraction reflects a student academic profile at the moment when he/she left the institution. Nevertheless, we consider here that it should be possible to reconstruct such a profile for any academic term in which the student was enrolled. This would involve defining some static (less variable) pieces of information in the student structure, and extending it to include metrics incrementally computed from the entrance of the student in the university until the considered term.  Next, numerical $X,Y$ pairs for both training and test sets can be created by a splitting approach applied on student structures. Here, $X$ represents a student feature matrix of dimension $n \times m$, with $n$ the number of students and $m$ the number of available features. $Y$ denotes a vector with $n$ numeric values, each one indicating  the enrollment status of a student in the database.  In the last step of the pipeline, machine learning methods are trained on the train set and evaluated on the test set. The best scored method is then output. As a final stage, the chosen method is trained over the whole dataset, and applied to predict the enrollment status for every enrolled student.


For the aim of this paper, we assume that any academic term can be expressed as a real number $yyyy.term$, with $yyyy$ its year of occurrence, and the fractional part $term$ a sequential value starting at 1 representing the order of that term in the year. From now on, we use the expressions ``date'', ``time'' and simply ``term''  in the text as synonyms for ``academic term''. Given a term $T$, $prev(T)$ and $next(T)$ are terms immediately before and after $T$.  

Let $I$ and $F$, $I\leq F$, be \textit{academic terms} mentioned by records in the database and corresponding to the initial and the final periods of study, respectively. Let $S$ be the set of \textit{student structures} created from the database for the students whose entrance to the educational institution happened in the time interval $[I..F]$. Note that there is no imposition yet on the exit terms, so $S$ may refer to students with exit term after $F$ or who are still enrolled in the institution. 

Given $I$, $F$ and $S$, we consider an inputted term $T$, with $I\leq T \leq F$, as a reference date for dropout prediction.  The prediction will be made for students in $S$ who were active at the beginning of term $T$. We also define the following subsets of $S$ that represent groups of students that exited the institution (by graduation or dropout) before $T$ or from $T$ on, or who were still enrolled: \textbf{$S^{I..prev(T)}$} contains the structures of students who exited in $[I..prev(T)]$;  \textbf{$S^{T..F}_T$}  contains the structures of students who exited in $[T..F]$ and with entrance term before or equal to $T$; and \textbf{$S^{\infty}_T$}  contains the structures of students who were still enrolled in the term $F$ and had entrance before or equal to $T$.

For any $s\in S$ representing a student, let $start(s)$ be his/her entrance term in the educational institution and $last(s)$ be the last academic term that appears in the academic records of this student. If the student has concluded the degree or dropped out, then $last(s)$ will be the last term in which him or her took a course or participated of any other academic activity. Let $end(s)$ be just one academic term ahead of $last(s)$, that is, $end(s)=next(last(s))$.  We define $x^t_s$, with $start(s)\leq t\leq end(s)$, as the numerical feature vectorized representation of $s$ at the beginning of the academic term $t$. The vector $x^t_s$ can have any characteristic attribute of the student that is constant. However, for time-based features like global average score and amount of enrolled terms, their appearance in $x^t_s$  should reflect the academic activities incrementally done only in the sequential period $[start(s)..prev(t)]$. For simplicity, we use the notation $x^{start}$, $x^t$ and $x^{end}$, respectively, for referring to the feature vectors of $s$ just after starting the degree (but before doing any course), at the very beginning of term $t$, and after doing all academic activities that the database has record of. Constructing $x^{start}$ demands extra definitions, since there may be no academic records (e.g., about courses taken and the enrollment time) for computing time-based features in this case\footnote{A possible way to define $x^{start}$ is to include the general static features and assume standard neutral values for all time-based features.}. Because of that, we do not consider the  start-term feature vectors in the evaluation section of the current work. A future study could approach dropout prediction for such freshman students.  

\begin{figure}[htbp]
    \centering
    \includegraphics[width=0.8\columnwidth, keepaspectratio]{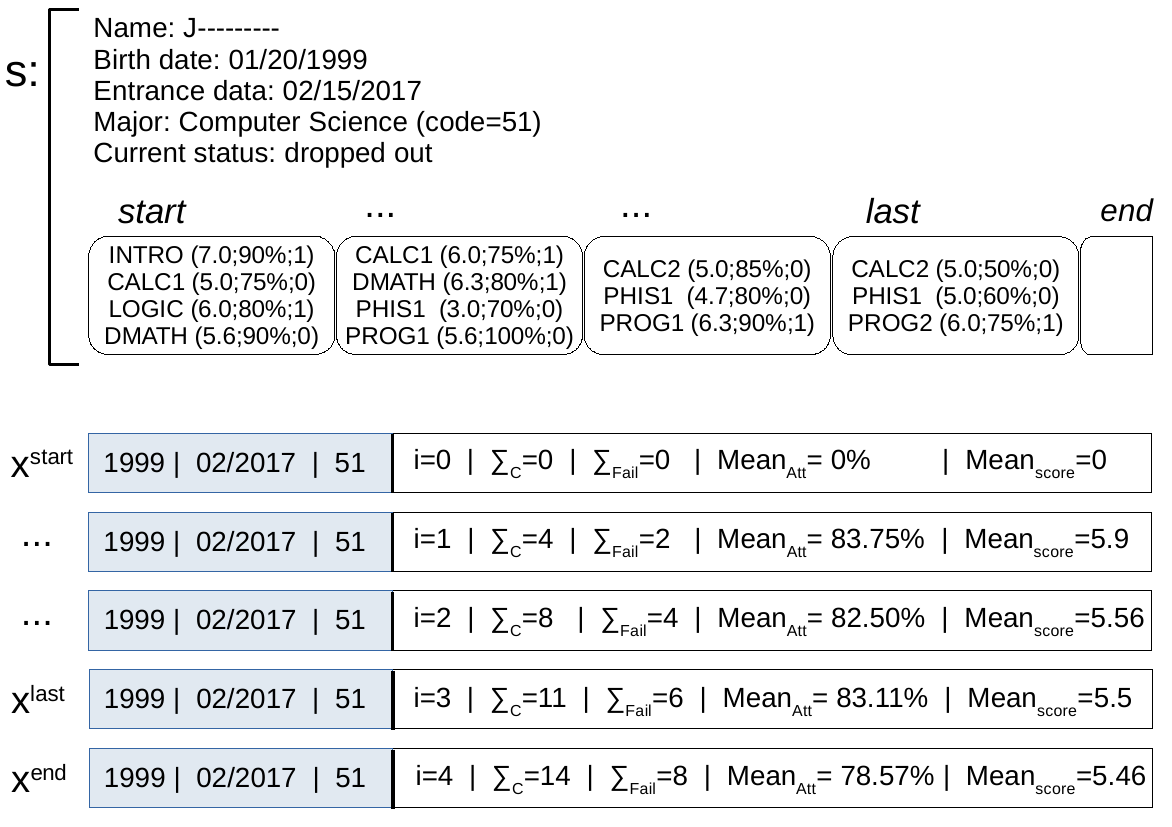}
    \caption{Exemplification of a student structure $s$ and of related feature vectors $x$ for incremental academic periods.} 
    \label{fig:s-x}
\end{figure}

Figure~\ref{fig:s-x} illustrates the concepts mentioned just above using an artificial and simple case.  The upper part of the figure exemplifies the structure $s$ for a student who entered in the educational institution for a particular major, took several courses and dropped out after four academic terms. The structure should hold information for each course taken by this student, such as the course code, his or her final score (from 0 to 10), the percentage of attendance in the course, and the final course result (which is 0 for failure and 1 for success). The lower part of the figure shows a feature vector $x$ for every incremental academic period $[{start}..t]$, with $t={start},\ldots,{end}$, generated from historical data available in $s$. We consider here a very compact feature-vector representation, with only eight features, just for illustration. The first three features (highlighted with darker background color) are static attributes of the student. We assume that they did not change during his/her academic degree and, therefore, keep them constant in all feature vectors built from $s$. The next five features are: the amount of completed academic terms  ($i$), the amount of  courses taken ($\sum_{C}$), the amount of non-succeeded courses ($\sum_{Fail}$, the mean of the attendance percentage in all courses ($Mean_{Att}$) and the student's global mean score ($Mean_{score})$. These latter features are based on historical data and need to be recalculated to include new information for each academic period, resulting in a corresponding $x$ vector.

In association to a feature vector $x^t_s$, we denote $y^t_s$ the enrollment status of the student in structure $s\in S$ in the academic term $t$. There are multiple ways of defining $y^t_s$. For the current work, as described in further in the evaluation section, we consider a binary dropout/conclusion classification and assume  $y^t_s$ to be invariant to $t$ and  equal to $0$ if the student dropped out or $1$ if he/she graduated. In our case, $y^t_s$ is not defined for students with the status ``enrolled''\footnote{Another possibility for defining the $y$ vector is using  multi-class classification with the three predefined status (``graduated'', ``dropout'' and ``enroll'`) or adopting a regression approach. Although interesting, the investigation of this aspect is outside the scope of the present work.}.


\section{Splitting Approaches}
\label{sec:splits}

We now present and discuss the splitting approaches. 
In general, they are related to a chosen academic term $T$. The idea is to train machine learning strategies on data coming from all cases in $S^{I..prev(T)}$, which refers to students who graduated or dropped out before $T$. Next, the trained models are used to predict  the dropout/conclusion status for all students who were active at the beginning of term $T$, specified by the set $S^{T..F}_T$. Then, the results of the models are compared against the actual status for those students.     

The first approach is the proportional splitting, like $70\%-30\%$ or $80\%-20\%$, commonly used in many other machine learning applications. This approach is named here just as ``Split A'', for simplicity. It consists of taking all related student structure data -- in this case, the union set $S_A=S^{I..prev(T)}\bigcup S^{T..F}_T$ -- and randomly splitting it into two subsets $S_A^{train}$ and $S_A^{test}$, for  training and test. Since no time information is employed in this split (students who left the institution at any term from $[I..F]$ can appear in either subsets), there is no clear guidance on which relative term $t$  to  use for constructing the feature vectors $x^t_s$, for each $s\in S_A$. Therefore, we adopt the most recent feature vectors from every student (represented by the feature vector $x^{end}$ in our notation),  as it would be done during a rough analysis of this matter\footnote{The vectors $x^{prev(end)}$ would be a much better proposition though, as described later in another splitting approach.}.  Besides, in order to allow a fair comparison to a temporal-split, we do not define a fixed proportion for the sizes of the training and test sets in \textit{Split A}. Instead, we make them to match the sizes of the two original sets that compose $S_A$, that is: $|S_A^{train}|=|S^{I..prev(T)}|$ and $|S_A^{test}|=|S^{T..F}_T|$ respectively.

It is important to mention that \textit{Split A} is not the appropriate approach for the prediction problem that we have at hand. It does not consider the natural time dependence between data available for training models and for prediction, what may imply in non-trustful performance results of the machine learning methods. In fact, applying this splitting resembles a much simpler problem of reconstructing momentarily-omitted enrollment status for some students randomly chosen, by learning the relationship between features and their associated status for the remaining students. 

The next splitting approaches are all temporal-based. The first one of them, called \textit{Split B1}, is a direct temporal version of \textit{Split A}, for which the training set is given by  $S^{I..prev(T)}$ and the test set comes from  $S^{T..F}_T$. The feature vectors $x^{end}$ are also employed for composing both training and test sets in this split.

\textit{Split B1} makes more sense than \textit{Split A}, since it emulates the situation of having previous data for training and then applying the resultant model to recent unknown cases. However, \textit{Split B1} does not fit the real problem yet. One obvious issue is that it still uses the most recent feature vector of every student in the test set. Thus, the prediction process is not estimating what will happen in a further term to a student who are ``theoreatically'' enrolled, but just rediscovers (again) the final status of a student who has already graduated or dropped out. To mitigate such an issue, we can change the test set for considering that the students are still enrolled, what can be accomplished by using not $x^{end}$ but $x^{last}$ as feature vectors for all student structures in the test set. Given that the training process should reflect the nature of the problem in the test step, the train set must also be constructed with $x^{last}$ vectors\footnote{Recall that $x^{last}$ is a numerical vector of features describing the academic situation of a student from the beginning of his/her degree to prior his/her last term.}.  This takes us to the next splitting approach.

\textit{Split B2} employs $S^{I..prev(T)}$ for training  and  $S^{T..F}_T$ for testing. It uses $x^{last}$ feature vectors for constructing both training and test sets.

In \textit{Split B2}, the temporal relation between training and test data sets is assured. Furthermore, the sense of using not complete academic data of the students but just what happened to them until their last terms is taken into consideration. Nevertheless, \textit{B2} still has a drawback: when predicting the status for a student who is active at the beginning of term $T$, academic information recorded after $T$ for such a student might be used in this splitting. This happens because $last(s)$ can be greater than $T$ for some $s\in S^{T..F}_T$.

\textit{Split B2T} avoids the above mentioned issue, by using in the test set only the academic information recorded until the beginning of term $T$. Formally, this split employs  $S^{I..prev(T)}$ for training  and  $S^{T..F}_T$ for testing as does \textit{Split B2}. It also considers $x^{last}$ feature vectors for constructing the training set. However, the test set consists now of $x^{T}_s$ for all $s\in S^{T..F}_T$.

\textit{Split B2T} is the first approach in the current paper that truly suits the prediction problem. It deals with all expected concerns and can be used just as presented. There is, however, one more aspect to consider that is inherent to the nature of the dropout prediction problem: the average number of enrolled terms for students who dropped out is smaller than that of students who graduated  (except for students who transferred from another degree and imported the scores of their taken courses). Consequently, the feature vector of students who are still in their first enrolled terms may seem very similar to that of students who dropped out early, which can confuse machine learning methods. As a way of compensating this effect, the next splitting approach augments the training set by including in it more samples of features vectors for students in early enrollment stages. 

\textit{Split B3T} uses  $S^{I..prev(T)}$ for training  and  $S^{T..F}_T$ for testing. The train set consists of $x^{t}_s$ for every $start(s)\leq t\leq last(s)$ and all $s\in S^{I..prev(T)}$. The test set consists of $x^{T}_s$ for all $s\in S^{T..F}_T$.

Finally, \textit{Split B4T} extends further the  training set by adding the final feature vector ($x^{end}$) of every student, that was discarded when defining the \textit{B2}, \textit{B2T} and \textit{B3T Splits}. Formally, it uses  $S^{I..prev(T)}$ for training  and  $S^{T..F}_T$ for testing. The training set consists of $x^{t}_s$ for every $start(s)\leq t\leq end(s)$ and all $s\in S^{I..prev(T)}$. The test set consists of $x^{T}_s$ for all $s\in S^{T..F}_T$.

Table~\ref{tab:splits} summarizes the main characteristics of the six splitting approaches, with the first one being the proportional split and the others temporal.

\begin{table}[htpb]
    \caption{Splitting approaches, with the feature vectors $x$ for the training and test sets.}
    \label{tab:splits}
   \centering
    \begin{tabular}{l|p{5cm}|p{2.3cm}}
         \textbf{Approach} & \textbf{Training set} & \textbf{Test set} \\ \hline
         $A$ - Proport. & $x^{end}_s, \forall s\in S_A^{train}$ & $x^{end}_s, \forall s\in S_A^{test}$ \\ 
         $B1$ - Temporal & $x^{end}_s, \forall s\in S^{I..prev(T)}$  & $x^{end}_s, \forall s\in S^{T..F}_T$ \\ 
         $B2$ - Temporal & $x^{last}_s, \forall s\in S^{I..prev(T)}$  & $x^{last}_s, \forall s\in S^{T..F}_T$ \\ 
         $B2T$ - Temporal & $x^{last}_s, \forall s\in S^{I..prev(T)}$  & $x^{T}_s, \forall s\in S^{T..F}_T$ \\
         $B3T$ - Temporal & $x^{t}_s, \forall t,s$ with $start(s)\leq t\leq last(s)$ and $s\in S^{I..prev(T)}$  & $x^{T}_s, \forall s\in S^{T..F}_T$ \\
        $B4T$ - Temporal & $x^{t}_s, \forall t,s$ with $start(s)\leq t\leq end(s)$ and  $s\in S^{I..prev(T)}$  & $x^{T}_s, \forall s\in S^{T..F}_T$ \\
    \end{tabular}
\end{table}

The above ways to split student data is not exhaustive. Other approaches are possible. For instance, the expansion of feature vectors done on the training sets in approaches \textit{B3T} and \textit{B4T} can be applied to the test set as well. In this case, the dropout prediction for a student will be given by the most frequent predicted answer in their feature vectors. Another way would be to group the feature vectors by the number of academic terms already attended, in order to train a ML model for each group, as carried out by Chen, Johri and Rangwala~\cite{Chen_Johri_Rangwala_2018}.

When several ML methods are available to be evaluated, it is necessary to run and compare them according to some performance measures. The method (or methods, for an ensemble) that proves to be most effective will be used for the final prediction of the dropout risk of students currently enrolled in the university. This involves retraining the method over the entire dataset, $S^{I..F}$, and applying it to a set of feature vectors constructed from $S^{\infty}_F$.

\section{Evaluation}
\label{sec:evaluation}

The data used in the experiments were gathered from a public university. It is composed of records from a total of 7.095 students from eight degrees in three academic faculties and of 242.417 course records. These are students who enrolled 
between the years of 2009 and 2019. After preprocessing the feature vector for each student consists of 27 features regarding demographic, social and academic attributes. 

The academic year consists of two terms each a semester long represented as `.1' and `.2'. There are in fact summer and winter terms which are one and a half months long, but due to their less frequency, the courses on these periods were mapped to the two main terms.

Dropout is defined when the student definitive exits from a course before its conclusion. Officially, it occurs when the student does not register in the following semester or he/she hits any of the university's exclusion criteria, such as exceeding the time limit for completing the course.

The test of dropout prediction for the different splitting approaches were executed for a period between I = 2009.1 and F = 2019.1 with the semesters T for when the prediction is made varying between 2012.2 and 2019.1 for active students at that time. The data was extracted during the 2019.2, and therefore we only have complete information for the previous semester, which is the last period we predict. For students before 2012.1 the rate of conclusion is very low because the regular conclusion time for most courses is 4 years. Therefore, students that concluded the degrees in before that period were only special cases. For maintaining a consistency in the comparison of A and B1 approaches the same proportion of train and test data were maintained.

Eight machine learning methods were used in each semester T for each splitting approach. For each period we identified the most effective and the second most effective methods. There are several approaches to determine effectiveness. Some of them can be: a simple accuracy mean for the period, the highest accuracy in the period etc. For the current work we used a point system. A method gains a point if it has the highest accuracy in two following semesters. The method with most points is chosen as the best, in case of a tie, we look at the mean for the whole period.   

\subsection{Results}
The accuracy results of all machine learning methods for each semester in all splitting approaches are presented in Table~\ref{tab:table-accuracy}. The table consists of six block of eleven lines each. Every block refers to a splitting approach. There is a line in all blocks for each method and a column for every academic term $T$. The resulting accuracy, measured in the range 0-100\%, of a method in an academic term is shown in a cell. The last line of each block has the mean accuracy of all methods for every semester. The most-right column has the mean accuracy of the methods for the whole period. The bottom-right corner of every block contains the mean accuracy for all methods and the considered semesters.

The best method in each splitting approach, determined using the previously cited point system, is displayed with an underline. The second best method is shown using italics. The highest accuracy value for in every semester is also displayed with an underline as well as the highest mean for the methods during the whole period.

\begin{table}[htpb]
 \caption{\label{tab:table-accuracy}Accuracy table for all methods in each splitting approach and all academic terms $T$ in $[2012.2..2019.1]$.}
    \centering
    \resizebox{\textwidth}{!}{%
        \begin{tabular}{lrrrrrrrrrrrrrrr}
        \toprule
        \multicolumn{16}{c}{\textbf{A} } \\ 
        \midrule
        \multicolumn{1}{c}{\textbf{Classifier} } & \multicolumn{1}{c}{\textbf{2012.2} } & \multicolumn{1}{c}{\textbf{2013.1} } & \multicolumn{1}{c}{\textbf{2013.2} } & \multicolumn{1}{c}{\textbf{2014.1} } & \multicolumn{1}{c}{\textbf{2014.2} } & \multicolumn{1}{c}{\textbf{2015.1} } & \multicolumn{1}{c}{\textbf{2015.2} } & \multicolumn{1}{c}{\textbf{2016.1} } & \multicolumn{1}{c}{\textbf{2016.2} } & \multicolumn{1}{c}{\textbf{2017.1} } & \multicolumn{1}{c}{\textbf{2017.2} } & \multicolumn{1}{c}{\textbf{2018.1} } & \multicolumn{1}{c}{\textbf{2018.2} } & \multicolumn{1}{c}{\textbf{2019.1} } & \multicolumn{1}{c}{\textbf{Mean} } \\ 
        \midrule
        Decision Tree & 98,29 & 98,15 & 99,25 & 99,27 & \uline{99,76}  & 99,50 & 99,45 & 99,19 & 99,59 & 99,78 & 99,16 & \uline{99,75}  & \uline{100,00}  & 98,72 & 99,28 \\
        Extra Trees & 98,88 & 99,32 & 99,25 & 99,59 & 99,53 & 99,80 & 99,56 & 99,36 & 99,59 & 99,78 & 99,53 & 99,62 & 99,83 & 98,72 & 99,45 \\
        Gradient Boosting & \uline{98,97}  & 98,74 & 99,25 & 99,45 & 99,67 & 99,80 & 99,62 & 99,25 & \uline{99,93}  & \uline{99,93}  & 99,53 & \uline{99,75}  & \uline{100,00}  & 98,72 & 99,47 \\
        KNN & 85,68 & 89,04 & 88,27 & 88,55 & 90,40 & 91,86 & 92,47 & 92,04 & 93,10 & 93,13 & 92,07 & 93,17 & 95,27 & 93,29 & 91,31 \\
        Naive Bayes & 95,41 & 95,72 & 95,65 & 94,35 & 96,01 & 95,73 & 95,85 & 95,00 & 95,31 & 94,85 & 94,40 & 94,05 & 96,85 & 96,81 & 95,43 \\
        \textit{Random Forest}  & 98,93 & 99,37 & 99,21 & \uline{99,64}  & 99,67 & 99,80 & 99,51 & \uline{99,42}  & \uline{99,93}  & 99,70 & \uline{99,63}  & \uline{99,75}  & \uline{100,00}  & \uline{99,04}  & 99,54 \\
        SVM & 51,22 & 52,68 & 53,65 & 49,75 & 50,55 & 51,61 & 52,76 & 55,67 & 55,28 & 57,28 & 57,18 & 55,70 & 59,37 & 59,74 & 54,46 \\
        \uline{XGBoost}  & 98,83 & \uline{99,73}  & \uline{99,44}  & \uline{99,64}  & 99,67 & \uline{99,85}  & \uline{99,73}  & \uline{99,42}  & \uline{99,93}  & 99,85 & \uline{99,63}  & \uline{99,75}  & \uline{100,00}  & \uline{99,04}  & \uline{99,61}  \\ 
        \midrule
        \multicolumn{1}{c}{\textbf{Mean} } & \textbf{90,77}  & \textbf{91,59}  & \textbf{91,75}  & \textbf{91,28}  & \textbf{91,91}  & \textbf{92,25}  & \textbf{92,37}  & \textbf{92,42}  & \textbf{92,83}  & \textbf{93,04}  & \textbf{92,64}  & \textbf{92,69}  & \textbf{93,91}  & \textbf{93,01}  & \textbf{92,32}  \\ 
        \bottomrule
        \multicolumn{16}{c}{\textbf{B1} } \\ 
        \toprule
        \multicolumn{1}{c}{\textbf{Classifier} } & \multicolumn{1}{c}{\textbf{2012.2} } & \multicolumn{1}{c}{\textbf{2013.1} } & \multicolumn{1}{c}{\textbf{2013.2} } & \multicolumn{1}{c}{\textbf{2014.1} } & \multicolumn{1}{c}{\textbf{2014.2} } & \multicolumn{1}{c}{\textbf{2015.1} } & \multicolumn{1}{c}{\textbf{2015.2} } & \multicolumn{1}{c}{\textbf{2016.1} } & \multicolumn{1}{c}{\textbf{2016.2} } & \multicolumn{1}{c}{\textbf{2017.1} } & \multicolumn{1}{c}{\textbf{2017.2} } & \multicolumn{1}{c}{\textbf{2018.1} } & \multicolumn{1}{c}{\textbf{2018.2} } & \multicolumn{1}{c}{\textbf{2019.1} } & \multicolumn{1}{c}{\textbf{Mean} } \\ 
        \midrule
        Decision Tree & 89,74 & 93,14 & 80,89 & 97,95 & 98,62 & 90,03 & 95,58 & 99,30 & 98,83 & \uline{99,40}  & 97,95 & 97,72 & 98,43 & \uline{98,40}  & 95,43 \\
        \textit{Extra Trees}  & \uline{99,41}  & 96,35 & 98,69 & \uline{99,59}  & \uline{99,48}  & \uline{99,55}  & 99,35 & 99,36 & 99,24 & 99,18 & 99,25 & 98,86 & 98,25 & 96,81 & \uline{98,81}  \\
        Gradient Boosting & 88,71 & 92,60 & 98,27 & 98,77 & 98,72 & 96,38 & 95,96 & 99,30 & 99,31 & 99,25 & \uline{99,44}  & 98,73 & 98,43 & 97,44 & 97,24 \\
        KNN & 41,25 & 84,30 & 85,09 & 85,64 & 85,42 & 85,56 & 85,93 & 86,23 & 86,47 & 87,53 & 87,13 & 85,82 & 84,97 & 80,83 & 82,30 \\
        Naive Bayes & 86,80 & 86,47 & 86,92 & 88,05 & 88,60 & 89,83 & 89,85 & 89,95 & 91,37 & 90,81 & 89,37 & 87,85 & 85,49 & 79,55 & 87,92 \\
        Random Forest & 98,93 & 94,90 & 98,36 & 99,45 & 99,34 & 99,26 & \uline{99,40}  & 99,30 & \uline{99,45}  & \uline{99,40}  & \uline{99,44}  & 98,86 & 98,25 & 97,13 & 98,68 \\
        SVM & 37,98 & 39,11 & 39,30 & 42,96 & 42,95 & 45,91 & 45,77 & 50,67 & 47,90 & 52,20 & 55,78 & 62,15 & 69,41 & 93,29 & 51,81 \\
        \uline{XGBoost}  & 93,99 & \uline{98,92}  & \uline{98,93}  & 98,86 & 98,81 & 99,21 & 99,13 & \uline{99,54}  & \uline{99,45}  & \uline{99,40}  & 99,16 & \uline{98,99}  & \uline{98,60}  & 97,44 & 98,60 \\ 
        \midrule
        \multicolumn{1}{c}{\textbf{Mean} } & \textbf{79,60}  & \textbf{85,72}  & \textbf{85,81}  & \textbf{88,91}  & \textbf{88,99}  & \textbf{88,21}  & \textbf{88,87}  & \textbf{90,46}  & \textbf{90,25}  & \textbf{90,90}  & \textbf{90,94}  & \textbf{91,12}  & \textbf{91,48}  & \textbf{92,61}  & \textbf{88,85}  \\ 
        \bottomrule
        \multicolumn{16}{c}{\textbf{B2} } \\ 
        \toprule
        \multicolumn{1}{c}{\textbf{Classifier} } & \multicolumn{1}{c}{\textbf{2012.2} } & \multicolumn{1}{c}{\textbf{2013.1} } & \multicolumn{1}{c}{\textbf{2013.2} } & \multicolumn{1}{c}{\textbf{2014.1} } & \multicolumn{1}{c}{\textbf{2014.2} } & \multicolumn{1}{c}{\textbf{2015.1} } & \multicolumn{1}{c}{\textbf{2015.2} } & \multicolumn{1}{c}{\textbf{2016.1} } & \multicolumn{1}{c}{\textbf{2016.2} } & \multicolumn{1}{c}{\textbf{2017.1} } & \multicolumn{1}{c}{\textbf{2017.2} } & \multicolumn{1}{c}{\textbf{2018.1} } & \multicolumn{1}{c}{\textbf{2018.2} } & \multicolumn{1}{c}{\textbf{2019.1} } & \multicolumn{1}{c}{\textbf{Mean} } \\ 
        \midrule
        Decision Tree & \uline{87,58}  & 92,14 & 87,65 & 89,79 & 89,93 & 92,01 & 91,05 & 90,81 & 92,55 & 90,81 & 90,95 & 90,51 & 89,16 & 83,33 & 89,88 \\
        \uline{Extra Trees}  & 86,11 & \uline{94,09}  & \uline{93,64}  & \uline{93,57}  & \uline{94,68}  & 94,19 & \uline{94,93}  & 94,01 & \uline{94,76}  & 94,55 & 94,12 & 92,79 & 90,73 & 85,26 & \uline{92,67}  \\
        Gradient Boosting & 49,88 & 90,93 & 90,74 & 90,70 & 91,50 & 91,76 & 93,51 & 92,85 & 93,58 & 93,58 & 93,47 & 93,29 & 90,21 & 84,62 & 88,61 \\
        KNN & 39,95 & 81,49 & 82,50 & 82,81 & 83,04 & 83,72 & 83,52 & 83,84 & 83,58 & 83,72 & 83,21 & 82,53 & 80,25 & 75,96 & 79,29 \\
        Naive Bayes & 83,62 & 83,21 & 84,37 & 85,50 & 86,13 & 87,49 & 87,89 & 87,73 & 89,30 & 88,50 & 87,22 & 85,19 & 82,87 & 76,92 & 85,42 \\
        Random Forest & 86,80 & 92,33 & 93,22 & 93,07 & 93,97 & 93,40 & 94,11 & 93,78 & 94,69 & 94,77 & 94,22 & 92,91 & 90,56 & 84,62 & 92,32 \\
        SVM & 37,95 & 39,05 & 39,21 & 42,96 & 42,95 & 45,91 & 45,77 & 50,64 & 47,90 & 52,20 & 55,78 & 62,15 & 69,58 & \uline{93,27}  & 51,81 \\
        \textit{XGBoost}  & 85,13 & 90,84 & 93,07 & 93,16 & 93,87 & \uline{94,84}  & 94,16 & \uline{94,24}  & 94,20 & \uline{94,92}  & \uline{94,40}  & \uline{93,80}  & \uline{90,91}  & 85,90 & 92,39 \\ 
        \midrule
        \multicolumn{1}{c}{\textbf{Mean} } & \textbf{69,63}  & \textbf{83,01}  & \textbf{83,05}  & \textbf{83,94}  & \textbf{84,51}  & \textbf{85,42}  & \textbf{85,62}  & \textbf{85,99}  & \textbf{86,32}  & \textbf{86,63}  & \textbf{86,67}  & \textbf{86,65}  & \textbf{85,53}  & \textbf{83,73}  & \textbf{84,05}  \\ 
        \bottomrule
        \multicolumn{16}{c}{\textbf{B2T} } \\ 
        \toprule
        \multicolumn{1}{c}{\textbf{Classifier} } & \multicolumn{1}{c}{\textbf{2012.2} } & \multicolumn{1}{c}{\textbf{2013.1} } & \multicolumn{1}{c}{\textbf{2013.2} } & \multicolumn{1}{c}{\textbf{2014.1} } & \multicolumn{1}{c}{\textbf{2014.2} } & \multicolumn{1}{c}{\textbf{2015.1} } & \multicolumn{1}{c}{\textbf{2015.2} } & \multicolumn{1}{c}{\textbf{2016.1} } & \multicolumn{1}{c}{\textbf{2016.2} } & \multicolumn{1}{c}{\textbf{2017.1} } & \multicolumn{1}{c}{\textbf{2017.2} } & \multicolumn{1}{c}{\textbf{2018.1} } & \multicolumn{1}{c}{\textbf{2018.2} } & \multicolumn{1}{c}{\textbf{2019.1} } & \multicolumn{1}{c}{\textbf{Mean} } \\ 
        \midrule
        Decision Tree & 65,17 & 65,86 & 60,77 & 60,51 & 62,57 & 69,81 & 74,30 & 75,76 & 79,46 & 84,35 & 88,12 & 90,24 & 89,11 & 82,41 & 74,89 \\
        \uline{Extra Trees}  & 58,40 & 64,56 & 65,78 & 66,00 & 69,15 & 70,77 & \uline{76,20}  & \uline{78,68}  & \uline{85,77}  & \uline{88,99}  & \uline{91,86}  & 92,55 & 90,54 & 83,79 & \uline{77,36}  \\
        Gradient Boosting & 53,70 & \uline{69,59}  & 65,39 & 63,15 & 64,91 & 64,33 & 70,56 & 75,05 & 80,10 & 84,44 & 89,46 & 90,79 & \uline{91,07}  & 83,10 & 74,69 \\
        KNN & 38,14 & 58,12 & 61,16 & 62,14 & 63,99 & 64,16 & 67,32 & 69,15 & 74,29 & 75,58 & 79,41 & 79,40 & 79,64 & 74,48 & 67,64 \\
        \textit{Naive Bayes}  & \uline{66,25}  & 67,75 & \uline{66,12}  & \uline{67,27}  & \uline{70,18}  & \uline{70,94}  & 75,53 & 78,10 & 83,50 & 84,19 & 85,92 & 84,15 & 83,04 & 75,17 & 75,58 \\
        Random Forest & 55,09 & 65,42 & 63,83 & 64,31 & 65,30 & 68,57 & 71,79 & 77,64 & 81,87 & 87,83 & 90,71 & 91,60 & 90,36 & 84,14 & 75,60 \\
        SVM & 36,90 & 36,47 & 38,07 & 40,29 & 42,01 & 42,45 & 44,47 & 45,50 & 46,60 & 48,43 & 54,60 & 59,49 & 68,93 & \uline{92,76}  & 49,78 \\
        XGBoost & 47,86 & 67,91 & 64,46 & 62,04 & 65,84 & 67,72 & 73,24 & 77,71 & 81,02 & 85,76 & 90,71 & \uline{92,95}  & \uline{91,07}  & 84,14 & 75,17 \\ 
        \midrule
        \multicolumn{1}{c}{\textbf{Mean} } & \textbf{52,69}  & \textbf{61,96}  & \textbf{60,70}  & \textbf{60,71}  & \textbf{62,99}  & \textbf{64,85}  & \textbf{69,18}  & \textbf{72,20}  & \textbf{76,58}  & \textbf{79,95}  & \textbf{83,85}  & \textbf{85,15}  & \textbf{85,47}  & \textbf{82,50}  & \textbf{71,34}  \\ 
        \bottomrule
        \multicolumn{16}{c}{\textbf{B3T} } \\ 
        \toprule
        \multicolumn{1}{c}{\textbf{Classifier} } & \multicolumn{1}{c}{\textbf{2012.2} } & \multicolumn{1}{c}{\textbf{2013.1} } & \multicolumn{1}{c}{\textbf{2013.2} } & \multicolumn{1}{c}{\textbf{2014.1} } & \multicolumn{1}{c}{\textbf{2014.2} } & \multicolumn{1}{c}{\textbf{2015.1} } & \multicolumn{1}{c}{\textbf{2015.2} } & \multicolumn{1}{c}{\textbf{2016.1} } & \multicolumn{1}{c}{\textbf{2016.2} } & \multicolumn{1}{c}{\textbf{2017.1} } & \multicolumn{1}{c}{\textbf{2017.2} } & \multicolumn{1}{c}{\textbf{2018.1} } & \multicolumn{1}{c}{\textbf{2018.2} } & \multicolumn{1}{c}{\textbf{2019.1} } & \multicolumn{1}{c}{\textbf{Mean} } \\ 
        \midrule
        Decision Tree & 60,26 & 54,71 & 47,69 & 51,16 & 61,99 & 70,49 & 75,03 & 80,17 & 63,74 & 86,01 & 87,26 & 84,42 & 78,93 & \uline{83,45}  & 70,38 \\
        \uline{Extra Trees}  & 52,61 & 63,96 & 63,83 & 73,07 & 76,37 & 78,63 & 79,27 & \uline{84,96}  & \uline{88,31}  & \uline{91,97}  & \uline{93,20}  & \uline{91,06}  & \uline{89,64}  & 82,07 & 79,21 \\
        Gradient Boosting & 41,09 & 65,86 & 58,44 & 62,41 & 64,38 & 72,64 & 73,30 & 81,66 & 81,09 & 89,07 & 92,53 & 89,84 & 88,04 & 82,07 & 74,46 \\
        KNN & 41,14 & \uline{71,21}  & 70,30 & 73,18 & 73,59 & 75,24 & 76,87 & 77,25 & 81,09 & 79,72 & 80,65 & 77,51 & 78,75 & 71,72 & 73,44 \\
        \textit{Naive Bayes}  & \uline{66,82}  & 70,02 & \uline{72,44}  & \uline{74,13}  & \uline{77,49}  & \uline{80,50}  & \uline{80,67}  & 82,31 & 87,47 & 86,34 & 86,97 & 85,23 & 83,39 & 75,86 & \uline{79,26}  \\
        Random Forest & 44,13 & 66,50 & 63,30 & 71,28 & 70,37 & 75,86 & 80,17 & 84,90 & 87,54 & 90,73 & 92,63 & 86,99 & 89,11 & 81,38 & 77,49 \\
        SVM & 36,90 & 37,72 & 40,64 & 61,04 & 57,99 & 58,62 & 55,36 & 54,50 & 53,40 & 49,50 & 45,40 & 40,52 & 31,07 & 6,90 & 44,97 \\
        XGBoost & 45,32 & 69,05 & 63,44 & 63,15 & 70,13 & 73,09 & 74,80 & 80,10 & 78,54 & 90,15 & 91,67 & 89,84 & 86,25 & 81,72 & 75,52 \\ 
        \midrule
        \multicolumn{1}{c}{\textbf{Mean} } & \textbf{48,53}  & \textbf{62,38}  & \textbf{60,01}  & \textbf{66,18}  & \textbf{69,04}  & \textbf{73,13}  & \textbf{74,43}  & \textbf{78,23}  & \textbf{77,65}  & \textbf{82,94}  & \textbf{83,79}  & \textbf{80,67}  & \textbf{78,15}  & \textbf{70,65}  & \textbf{71,84}  \\ 
        \bottomrule
         \multicolumn{16}{c}{\textbf{B4T} } \\ 
        \toprule
        \multicolumn{1}{c}{\textbf{Classifier} } & \multicolumn{1}{c}{\textbf{2012.2} } & \multicolumn{1}{c}{\textbf{2013.1} } & \multicolumn{1}{c}{\textbf{2013.2} } & \multicolumn{1}{c}{\textbf{2014.1} } & \multicolumn{1}{c}{\textbf{2014.2} } & \multicolumn{1}{c}{\textbf{2015.1} } & \multicolumn{1}{c}{\textbf{2015.2} } & \multicolumn{1}{c}{\textbf{2016.1} } & \multicolumn{1}{c}{\textbf{2016.2} } & \multicolumn{1}{c}{\textbf{2017.1} } & \multicolumn{1}{c}{\textbf{2017.2} } & \multicolumn{1}{c}{\textbf{2018.1} } & \multicolumn{1}{c}{\textbf{2018.2} } & \multicolumn{1}{c}{\textbf{2019.1} } & \multicolumn{1}{c}{\textbf{Mean} } \\ 
        \midrule
        Decision Tree & 50,03 & 58,60 & 48,71 & 66,95 & 67,69 & 71,74 & 77,82 & 80,36 & 83,64 & 84,27 & 87,17 & 81,98 & 81,96 & 81,72 & 73,05 \\
        \textit{Extra Trees}  & 50,54 & 65,69 & 65,29 & \uline{75,29} & 75,44 & 77,50 & 78,94 & 85,29 & 87,39 & \uline{92,80} & \uline{93,20} & 90,65 & \uline{90,00} & 81,38 & 79,24\\
        Gradient Boosting & 40,88 & 67,80 & 59,84 & 64,31 & 67,30 & 72,36 & 73,18 & 84,45 & 84,77 & 89,57 & 92,72 & \uline{90,79} & 89,11 & 82,07 & 75,65 \\
        KNN & 44,91 & \uline{72,02} & 70,69 & 72,97 & 73,93 & 75,07 & 77,10 & 77,51 & 82,08 & 80,30 & 81,32 & 78,86 & 79,82 & 72,76 & 74,24 \\
        \uline{Naive Bayes}  & \uline{67,49} & 70,02 & \uline{71,71} & 74,29 & 77,19 & \uline{80,22} & \uline{80,62} & 82,50 & \uline{87,75} & 86,51 & 86,97 & 84,96 & 83,39 & 75,86 & \uline{79,25}  \\
        Random Forest & 48,89 & 69,64 & 66,55 & 72,70 & \uline{77,97} & 75,92 & 80,17 & \uline{85,55} & \uline{87,75} & 89,57 & 93,01 & 89,70 & 88,57 & \uline{82,76} & 79,20 \\
        SVM & 36,90 & 37,72 & 39,48 & 56,76 & 54,97 & 58,62 & 55,36 & 54,50 & 53,40 & 49,50 & 45,40 & 40,52 & 31,07 & 6,90 & 44,36 \\
        XGBoost & 42,64 & 70,24 & 64,56 & 67,42 & 70,61 & 72,98 & 75,42 & 78,55 & 82,15 & 89,49 & 91,67 & 88,75 & 84,29 & 81,72 & 75,75 \\ 
        \midrule
        \multicolumn{1}{c}{\textbf{Mean} } & \textbf{47,78} & \textbf{63,97} & \textbf{60,85} & \textbf{68,84} & \textbf{70,64} & \textbf{73,05} & \textbf{74,83} & \textbf{78,59} & \textbf{81,12} & \textbf{82,75} & \textbf{83,93} & \textbf{80,78} & \textbf{78,53} & \textbf{70,65} & \textbf{72,59}  \\ 
        \bottomrule
        \end{tabular}
    }\\
\end{table}

\begin{table}[htpb]
   \caption{\label{tab:table-students} Quantity of students for each train and test sets and all academic terms $T$ in $[2012.2..2019.1]$. The last line of the table shows the number of students with status ``enrolled'' in the database (given by $|S^{\infty}_T|$)  for each term.}
    \centering
    \resizebox{\textwidth}{!}{%
        \begin{tabular}{lcccccccccccccc} 
        \toprule
         & \textbf{2012.2}  & \textbf{2013.1}  & \textbf{2013.2}  & \textbf{2014.1}  & \textbf{2014.2}  & \textbf{2015.1}  & \textbf{2015.2}  & \textbf{2016.1}  & \textbf{2016.2}  & \textbf{2017.1}  & \textbf{2017.2}  & \textbf{2018.1}  & \textbf{2018.2}  & \textbf{2019.1}  \\ 
        \midrule
        \textbf{A / B1 Train}  & 380 & 578 & 738 & 984 & 1125 & 1461 & 1686 & 1976 & 2285 & 2526 & 2821 & 3155 & 3385 & 3667 \\
        \textbf{A / B1 Test}  & 2046 & 2217 & 2140 & 2193 & 2105 & 2015 & 1833 & 1721 & 1449 & 1339 & 1072 & 790 & 572 & 313 \\
        \textbf{B2 Train}  & 373 & 570 & 728 & 971 & 1112 & 1448 & 1673 & 1963 & 2271 & 2512 & 2807 & 3141 & 3371 & 3653 \\
        \textbf{B2 Test}  & 2045 & 2215 & 2137 & 2193 & 2105 & 2015 & 1833 & 1720 & 1449 & 1339 & 1072 & 790 & 572 & 312 \\
        \textbf{B2T / B3T / B4T Train}  & 373 & 570 & 728 & 971 & 1112 & 1448 & 1673 & 1963 & 2271 & 2512 & 2807 & 3141 & 3371 & 3653 \\
        \textbf{B2T / B3T / B4T Test}  & 1935 & 1848 & 2057 & 1894 & 2052 & 1769 & 1790 & 1543 & 1412 & 1208 & 1044 & 738 & 560 & 290 \\ 
        \midrule
        \textbf{Enrolled}  & 153 & 183 & 309 & 377 & 561 & 624 & 871 & 949 & 1295 & 1386 & 1786 & 1899 & 2332 & 2462 \\
        \bottomrule
        \end{tabular}
    }\\
\end{table}

As an overall trend, we can see that the accuracy values increase over time in general. This can be due to two factors. First, the increase in size of the training data tends to improve the performance of the machine learning methods. Second, the size of the test cases starts to decrease considerably by $T=2015.1$ and the test set gets more and more specialized. This second factor happens because the number of students with enrollment status equals to ``graduated'' or ``dropout'' (and for whom we can validate the prediction) decreases as the term $T$ gets closet to $F$, while the number of enrolled students ($|S^{\infty}_T|$) increases. Table~\ref{tab:table-students} shows these changes in the size of the sets, for all terms $T$ and splitting approaches. In the last semester considered for analysis ($T=2019.1$), there are only 290 to 313 students for whom we can validate the dropout prediction\footnote{The small change in the number of students in training and test sets from one splitting approach to another is due to the option of using or not their last academic term in the prediction task. As we explained in Section~\ref{sec:splits}, some splitting approaches do not use information about the students' last academic term or related to the term $T$. Thus, students that had just started a degree or had completed only one semester are not considered in those approaches. We also note that it could be possible to retain such students in the analysis by including their $x^start$ feature vectors. However, as also discussed before, this would need some extra definitions, and we opted not to include them in the current work since the number of such cases was small.}.  These students left the educational institution in 2019.1 or at the beginning of 2019.2. All other students (2462) were still pursuing their degree. We realized, by an analysis of the results in a semester $T$, that the prediction accuracy rate decreases as the student's semester of departure from the institution moves away from $T$. We can also see in the table that, there is a drop in accuracy in almost all methods and approaches after 2017.2. This is consistent with a curriculum change for some degrees that happened in the university between 2016 and 2017.  Very likely, this historical event effected the student profiles and led to the lower performance of the methods. Another general trend, related to the dataset itself, is that the number of dropout cases becomes  relatively larger than the number of graduated cases after $2017.1$. 

We now discuss and compare the splitting approaches. The \textit{Split A}, what is not a realistic model to the problem, has the overall best accuracies out of all other splittings. Its results are already very high at the first academic term and increase slowly but in a stable way. Since this splitting approach uses a mixture of information from past and present students, it makes the methods perform better and less sensitive to academic changes in the institution. The XGBoost was the best method in \textit{Split A}, followed by the Random Forest.

\textit{Split B1} separates past and present students respectively into training and test data. This leads to a reduction in the overall accuracy (that can be seen by the mean values), but the results are still very high for some methods.  The XGBoost was the best method in \textit{Split B1}, and the Extra Trees was the second best. We recall that \textit{B1} utilizes the complete data for the students until the moment they leave the university, which is also not present in a real setting.   

In \textit{Split B2}, we do not use the students' last semester for training and testing. As a consequence of this, the performance is lower of that reached by the \textit{B1} approach. With this split, The Extra Trees was the best method, followed by the XGBoost.

Aiming at solving a last issue, \textit{Splits B2T}, \textit{B3T} and \textit{B4T} only employ information for feature vectors that is available before the semester $T$ of prediction, therefore being suitable approaches to the problem. Since we use less information than the previous approaches, the accuracy rates reached by the machine learning methods in these splitting approaches are much lower than in $B2$, particularly for the first academic terms.  \textit{Splits B3T} and \textit{B4T} gradually improved the results over \textit{B2T}, with the later splitting been the best one. Extra Trees and Naive Bayes were the best methods in these splittings, with the Naive Bayes showing a higher accuracy at the first academic terms and the Extra Trees in the last semesters.  

\begin{figure}[hptb]
    \centering
    \includegraphics[width=1\columnwidth, keepaspectratio]{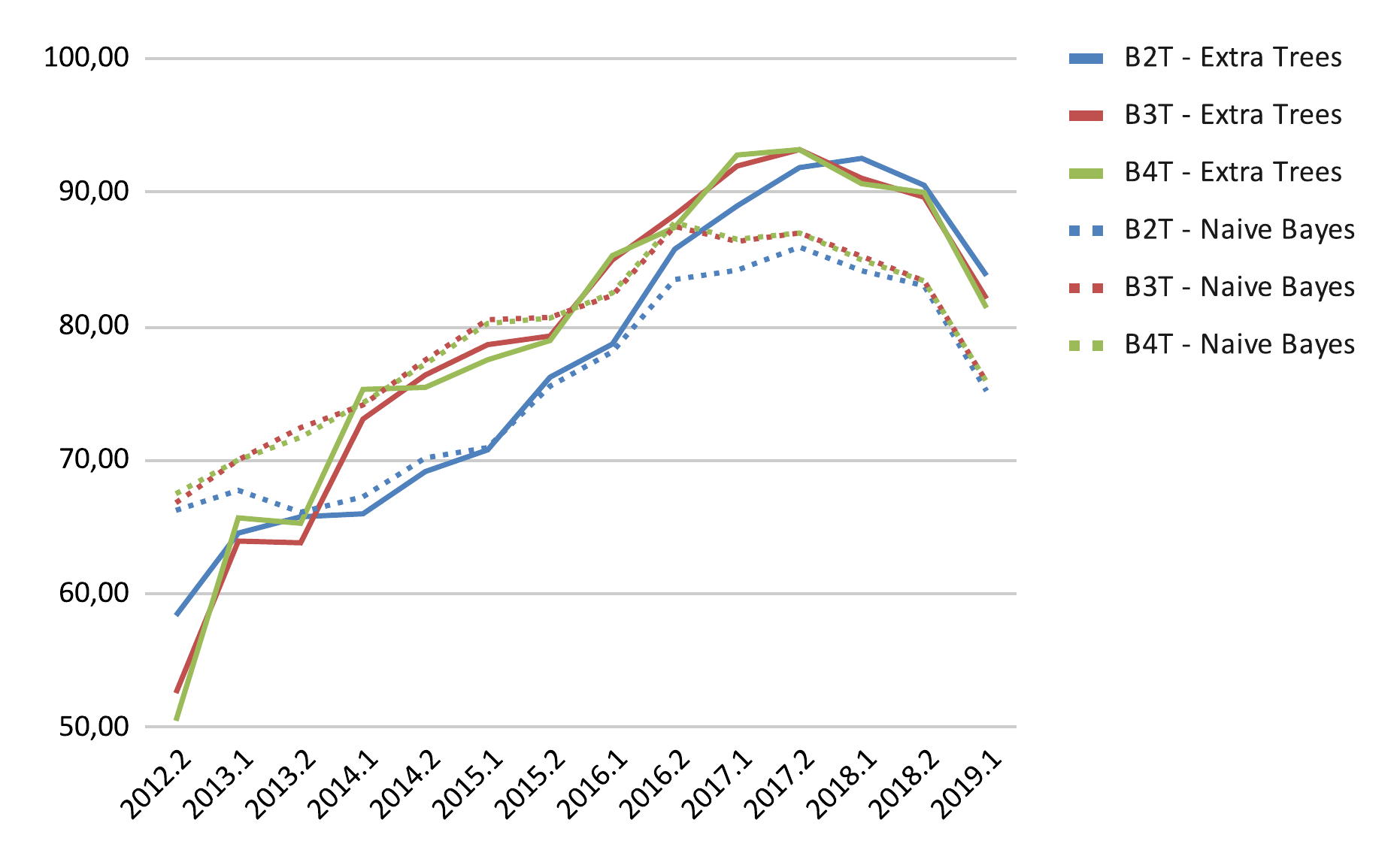}
    \caption{Accuracy for Extra Trees and Naive Bayes for \textit{B2T}, \textit{B3T} and \textit{B4T}.} 
    \label{fig:chart-both}
\end{figure}

Comparing the results of all six splitting approaches, one can see that the selected splitting impacts on the choice of which method to use in a following semester. Overall, three methods stood out from the rest as having the best performance: Extra Trees, XGBoost and Naive Bayes. Since XGBoost yielded better results only for unrealistic settings, we are not considering it in our next analysis. Figure~\ref{fig:chart-both} shows a comparison between the Extra Trees and the Naive Bayes only for the \textit{B2T},  \textit{B3T} and  \textit{B4T}. It shows that  \textit{B3T} and  \textit{B4T} splitting approaches yield better results overall with the latter having a slightly higher accuracy over time. The methods show a better performance in different periods of time. Therefore, a combination of them would possibly yield a more robust solution.

\begin{figure}[h!]
    \centering
    \includegraphics[width=0.85\textwidth, keepaspectratio]{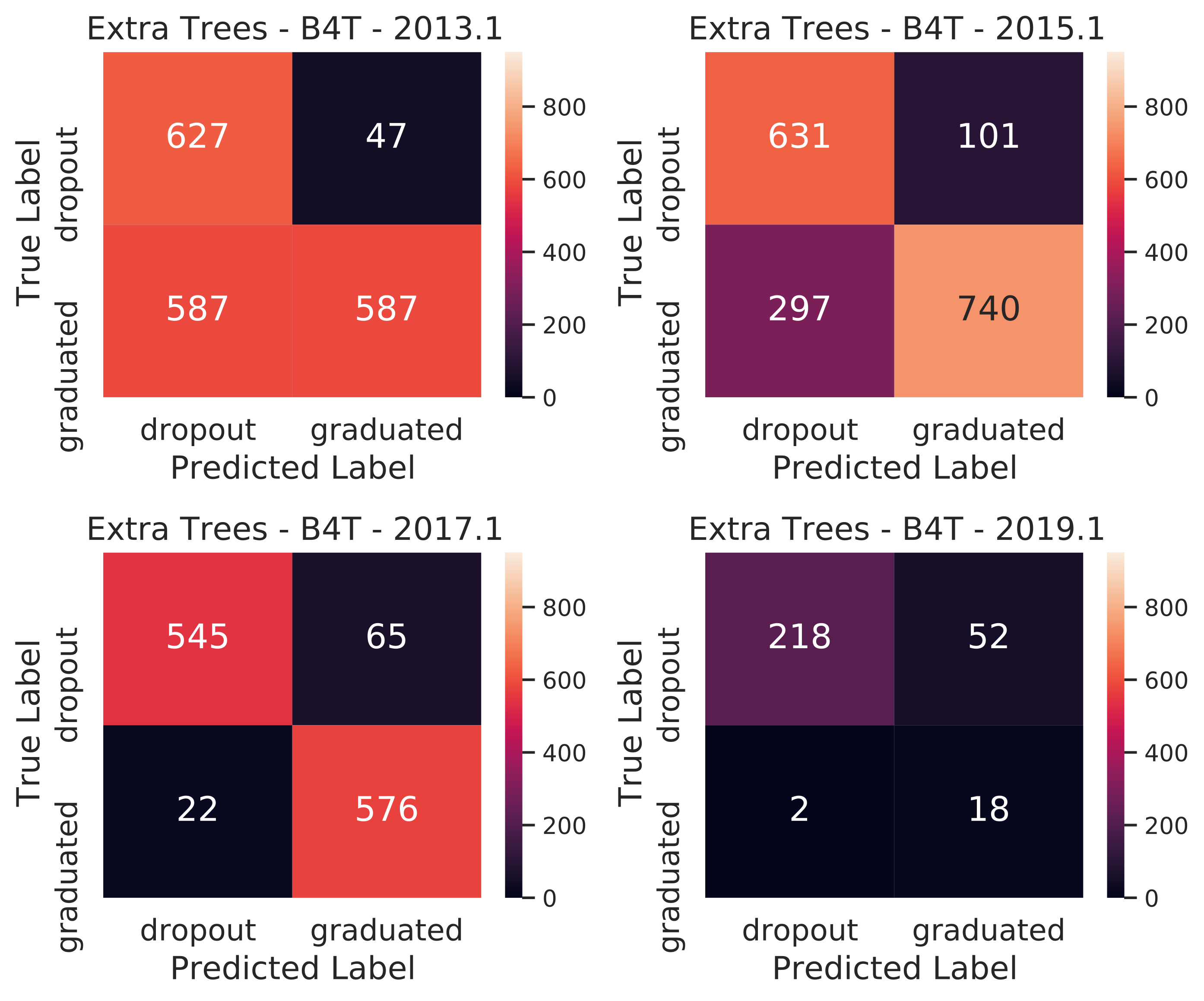} \\
    \vspace{0.1in}
     \includegraphics[width=0.85\textwidth, keepaspectratio]{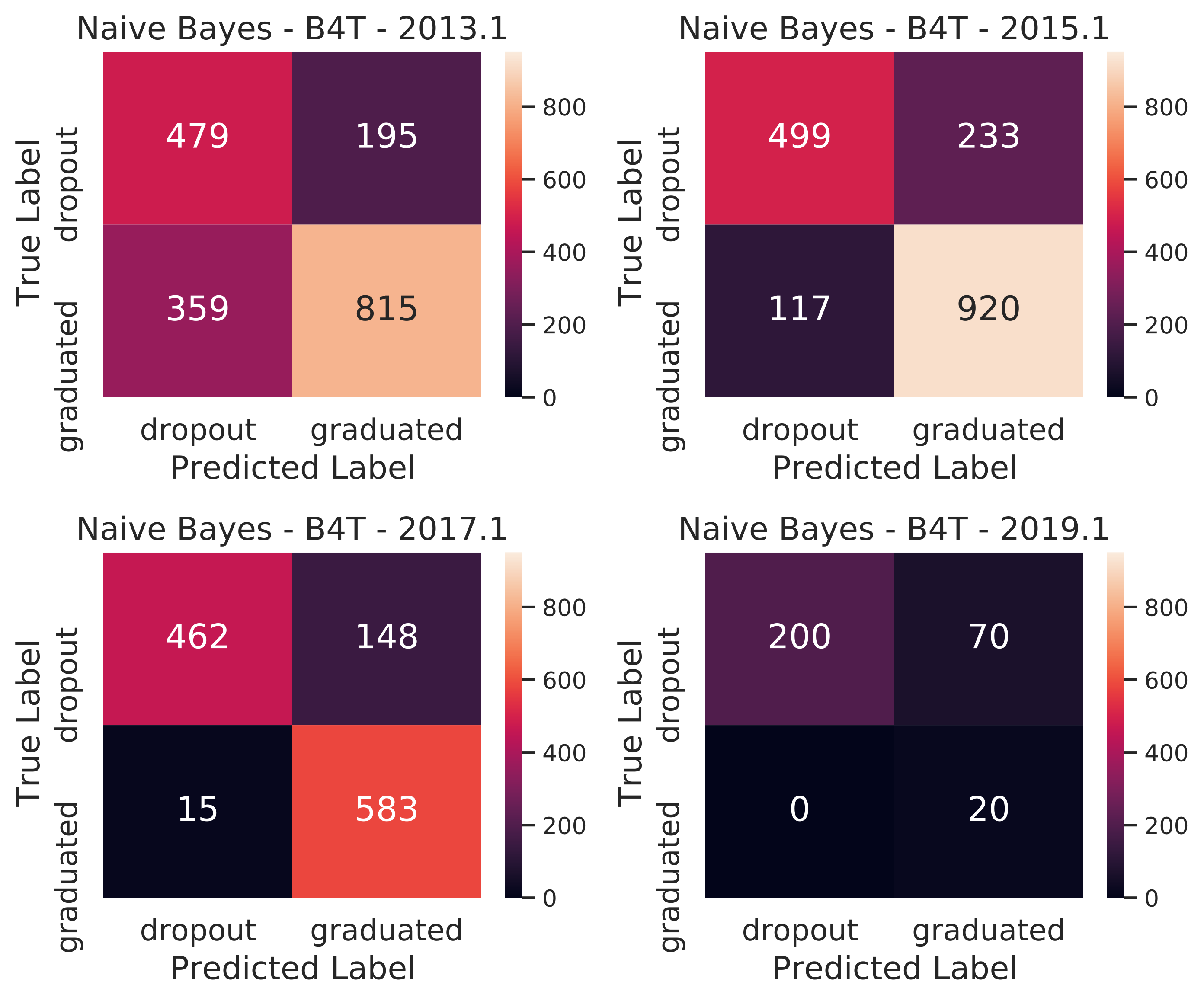}
    \caption{Confusion matrices for the Extra Trees and the Naive Bayes methods in the \textit{B4T Split}.} 
    \label{fig:cm_B4T_confusionmatrices}
\end{figure}

As a final evaluation of the results that goes beyond the accuracy measurement, we present the confusion matrices for the Extra Trees and the Naive Bayes methods with the \textit{B4T} splitting approach in Figure~\ref{fig:cm_B4T_confusionmatrices}, for four years (2013.1, 2015.1, 2017.1 and 2019.1).

\section{Conclusions}
\label{sec:conclusion}

In this paper, we formalize and compare approaches to split the data into training and test sets for the prediction of dropout in higher education using machine learning methods. A proportional splitting approach, which is commonly used in various application domains and even appears in some scientific papers about prediction in the academic context, is not suitable for the real problem under study. It is the temporal splitting that correctly represents this prediction scenario. Still, some details need to be considered, such as the discarding of information that is outside the period under analysis, as well as the construction of multiple feature vectors for the same student, with each vector representing the incremental situation of a student up to a given academic term.

In an experiment conducted with data from a public higher education institution, the \textit{B4T} temporal splitting approach showed better results. Two machine learning methods stood out when tackling the real problem, each of them being most successful at different times. When analyzing the mean accuracy of the temporal splittings \textit{B2T} to \textit{B4T},  we see that the values are generally below 90\%, which demonstrates that there is still space for improvement through scientific research. 
 
As future work, we suggest extending the study to include other forms of temporal splitting. It would also be interesting to develop a more effective strategy to determine which of the methods to use for the prediction of dropout risk for students currently enrolled. Moreover, the discussions and formalization of the temporal splits presented here are general enough to suit other prediction tasks in education, such as for academic performance prediction. They can even be useful in more distant application domains, given that there is a growing dataset based on historical sequences, with new sequential data available for feeding a prediction task and previous sequences for composing the training corpus. For instance, recommendation systems in streaming platforms rely on this type of dataset and can benefit from our discussions.      
 
\section*{Acknowledgment}
This study was financed in part by the Coordenação de Aperfeiçoamento de Pessoal de Nível Superior – Brasil (CAPES) – Finance Code 001. 





\bibliographystyle{unsrtnat}
 \bibliography{references}  






\end{document}